%% file: main.tex
\documentclass[sigconf,nonacm]{acmart}

\usepackage[]{multibib}
\newcites{BI}{Appendix: References over page limit}
\newcites{reg}{References}

\AtBeginDocument{%
  \providecommand\BibTeX{{%
    \normalfont B\kern-0.5em{\scshape i\kern-0.25em b}\kern-0.8em\TeX}}}

\usepackage{todonotes}
\usepackage{verbatim}
\usepackage[normalem]{ulem}
\usepackage{tabularx}
\usepackage{adjustbox}
\usepackage{booktabs}
\usepackage{ragged2e}
\usepackage{enumitem}
\usepackage{array,tikz}

\usetikzlibrary{tikzmark,arrows.meta}

\newcolumntype{P}[1]{>{\RaggedRight\hspace{0pt}}p{#1}}

\def\tikzmarkhf#1{\hspace{\fill}\tikzmark{#1}}

\setcopyright{none}
\renewcommand\footnotetextcopyrightpermission[1]{} 

\usepackage{soul}
\definecolor{mypink1}{RGB}{255, 255, 255}
\definecolor{mygreen1}{RGB}{255, 255, 255}
\definecolor{myblue1}{RGB}{255, 255, 255}

\copyrightyear{2022}
\acmYear{2022}
\setcopyright{acmlicensed}\acmConference[AIES'22]{Proceedings of the 2022 AAAI/ACM Conference on AI, Ethics, and Society}{August 1--3, 2022}{Oxford, United Kingdom}
\acmBooktitle{Proceedings of the 2022 AAAI/ACM Conference on AI, Ethics, and Society (AIES'22), August 1--3, 2022, Oxford, United Kingdom}
\acmPrice{15.00}
\acmDOI{10.1145/3514094.3534196}
\acmISBN{978-1-4503-9247-1/22/08}

\begin{document}

\title{The worst of both worlds: A comparative analysis of errors in learning from data in psychology and machine learning}


\newcommand{\tsc}[1]{\textsuperscript{#1}} 
\author{Jessica Hullman\tsc{1,4}, Sayash Kapoor\tsc{2}, Priyanka Nanayakkara\tsc{1}, Andrew Gelman\tsc{3}, Arvind Narayanan\tsc{2}}
\authornote{First author is corresponding author.\\1. Northwestern University. $\{$jhullman@, priyankan@u.$\}$northwestern.edu\\2. Princeton University. $\{$sayashk@, arvindn@cs.$\}$princeton.edu\\3. Columbia University. gelman@stat.columbia.edu\\4. Microsoft Research New York City. }









\renewcommand{\shortauthors}{J. Hullman, S. Kapoor, P. Nanayakkara, A. Gelman, and A. Narayanan}

\newcounter{daggerfootnote}
\newcommand*{\daggerfootnote}[1]{%
    \setcounter{daggerfootnote}{\value{footnote}}%
    \renewcommand*{\thefootnote}{$\ddag$}%
    \footnote[2]{#1}%
    \setcounter{footnote}{\value{daggerfootnote}}%
    \renewcommand*{\thefootnote}{\arabic{footnote}}%
    }

\begin{abstract}
Arguments that machine learning (ML) is facing a reproducibility and replication crisis suggest that some published claims in ML research cannot be taken at face value. These concerns inspire analogies to the replication crisis affecting the social and medical sciences. They also inspire calls for the integration of statistical approaches to causal inference and predictive modeling. A deeper understanding of what reproducibility concerns in supervised ML research have in common with the replication crisis in experimental science puts the new concerns in perspective, and helps researchers avoid ``the worst of both worlds,'' where ML researchers begin borrowing methodologies from explanatory modeling without understanding their limitations and vice versa. We contribute a comparative analysis of concerns about
inductive learning that arise in causal attribution as exemplified in psychology versus predictive modeling as exemplified in ML. We identify themes that re-occur in reform discussions, like overreliance on asymptotic theory and non-credible beliefs about real-world data generating processes. We argue that in both fields, claims from learning are implied to generalize outside the specific environment studied (e.g., the input dataset or subject sample, modeling implementation, etc.) but are often difficult to refute due to underspecification of key parts of the learning pipeline. In particular, errors being acknowledged in ML expose cracks in long-held beliefs that optimizing predictive accuracy using huge datasets absolves one from having to consider a true data generating process or formally represent uncertainty in performance claims. We conclude by discussing risks that arise when
sources of errors are misdiagnosed and the need to acknowledge the role of human inductive biases in learning and reform.

\end{abstract}



\settopmatter{printfolios=true}

\maketitle

\input{0_intro}

\input{1_background}
\input{2_1_data}
\input{2_2_model}
\input{2_3_objective_etc}

\input{2_4_communication}

\input{3_discussion}

\vspace{-2mm}
\section{Acknowledgments}
We thank Jake Hofman, Cyril Zhang, Jean Czerlinski Oretga for comments. Hullman is supported by the NSF (IIS-1930642) and a Microsoft Faculty Fellowship, Narayanan by the NSF (IIS-1763642), and Gelman by the ONR.

\bibliographystyle{ACM-Reference-Format}
\typeout{}
\balance
\bibliography{ref.bib}

\end{document}

%% file: 0_intro.tex
\vspace{-2mm}
\section{Introduction}

The replication crisis in psychology and the social and medical sciences has spread to a general concern about scientific claims that are based on statistical significance. Similar attention has recently been drawn to replication challenges regarding empirical claims in artificial intelligence (AI) and machine learning (ML). There are direct concerns about {\em reproducibility}---published results cannot be reproduced using the same software and data due to unavailable tuning parameters, random seeds, and other configuration settings or computational infrastructure that are not available to outsiders---{\em replication}---where re-implementing described methods does not produce the same results due to unacknowledged dependencies, such as specific implementations, and---{\em generalizability or robustness}---where methods may work well under certain conditions but fail when applied to new problems or in the world~\cite{pineau2021improving}, where vulnerability to adversarial manipulations may be costly. For example, the identification of examples by which computer vision models could be tricked into misclassification by manipulations not visible to the human eye~\cite{szegedy2013intriguing} has inspired subsequent research proposing a variety of explanations for the apparent brittleness of performance (e.g.,~\cite{d2020underspecification,goodfellow2014explaining,geirhos2020shortcut}). Terms like ``alchemy''~\cite{hutson2018has} and ``graduate student descent'' are used to describe how researchers combine optimizations to often opaque parameters to achieve performance benchmarks. Model performance evaluations are conducted without acknowledging sources of error~\cite{agarwal2021deep,gorman2019we,liao2021we} and can involve data filtering decisions that impact achievable accuracy~\cite{bone2015applying,liaoforward,liao2021we}. 


Some amount of replication failure is inevitable: the nature of empirical research is to try out ideas that may work in some settings but not others. When claims are published, uncertainty about generalizability is inherent.
However, once systemic problems are recognized, corrective actions should be taken, and claims discounted---especially when they cannot be externally reproduced \cite{gelman2014beyond,button2013power,ioannidis2008}. 

One way that authors call attention to concerns in ML research is analogizing them to the replication crisis in psychology~\cite{bell2021perspectives,bouthillier2020survey,hofman2021integrating,heaven2020ai}. While psychologists discussed fundamental issues with conventional approaches to inference as early as the 1960s~\cite{meehl1967theory, cohen1992statistical}, in the last decade critics brought concerns to the forefront, demonstrating how motivated researchers can obtain false positives under various conditions~\cite{simmons2011false,francis2012a,francis2012b} and that many published conclusions about human behavior in psychology research cannot be replicated~\cite{nosek2015,dreber2015}. 
These revelations spur hard questions about what are necessary conditions for science, how to resolve uncertainty about published claims, and how to shift incentives. 

Despite their focus on predictive modeling (and abundant recent successes in terms of performance on benchmarks, e.g.,~\cite{eckersley2017eff} and adoption in real-world applications, e.g.,~\cite{magoulas2020ai}), fields like AI and ML could learn from psychology's ongoing attempts to diagnose sources of non-replicability and reform conventional use and reporting of methods in the causally-focused explanatory modeling paradigm prevalent in psychology.\footnote{Also known as attribution~\cite{efron2020prediction}, and typically involving estimation of regression surfaces and assignment of significance to individual predictors.} 
Taking a wider perspective on learning failures is well aligned with the idea of integrative modeling, referring to approaches that combine aspects of both paradigms~\cite{hofman2017prediction,hofman2021integrating,yarkoni2017choosing,rocca2021putting}. For example, social scientists might use prediction along with explanation to reduce overfitting to noisy experiment results, while researchers in fields like ML can incorporate explanatory methods to ascertain what a model appears to have learned. Integrative modeling acknowledges how researchers frequently misunderstand the relationship between explanation and prediction, assuming, e.g., that  
models that succeed in explaining have greater predictive validity~\cite{yarkoni2017choosing} than those that appear less psychologically plausible (\cite{hagerty1991comparing,shmueli2010explain} as cited in \cite{yarkoni2017choosing}) or that a model that achieves high predictive accuracy won't deviate much from what a human considers to be a plausible decision rule~\cite{geirhos2020shortcut}. 
But to avoid integrative modeling leading to ``the worst of both worlds,'' researchers will need to understand subtle differences in ways in which inferences can be limited in each paradigm.
To date, connections that authors have drawn between these two reform discussions have been piecemeal, leaving it unclear what lessons, if any, might be gained from putting these domains in conversation.

To address this gap, we contribute a detailed comparison of limitations of inference in causally-driven explanatory versus predictive modeling. 
Our analysis is synthesized from formal and informal arguments made in hundreds of papers we collected through online search, citation tracing, and our involvement in events and scholarly discussions on replication and reproducibility over multiple years.
While we surface issues that affect various areas in psychology and ML, we ground our discussion around examples from experimental social psychology, which like ML relies on data reflecting human behavior and uses controlled comparisons to produce claims, and empirical research in supervised discriminative learning (i.e., classification) methods, including deep neural nets (DNNs) that encapsulate many recent concerns.

Our results highlight where concerns across the two domains can stem from similar types of oversights, including overreliance on theory, underspecification of learning goals, non-credible beliefs about real-world data generating processes, overconfidence based in conventional faith in certain procedures (e.g., randomization, test-train splits), and tendencies to reason dichotomously about empirical results. In both fields, claims from learning are implied to generalize outside the specific environment studied (e.g., the input dataset or subject sample, modeling implementation, etc.) but are often impossible to refute due to undisclosed sources of variance in the learning pipeline. We argue in particular that many of the errors recently discussed in ML expose the cracks in long-held beliefs that optimizing predictive accuracy using huge datasets absolves one from having to consider a true data generating process or formally represent uncertainty in performance claims.
At the same time, the goals of ML are inherently oriented toward addressing learning failures, suggesting that lessons about irreproducibility could be resolved through further methodological innovation in a way that seems unlikely in social psychology. This assumes, however, that ML researchers take concerns seriously and avoid overconfidence in attempts to reform. 
We conclude by discussing risks that arise when sources of errors are misdiagnosed and the need to acknowledge the role that human inductive biases play in learning and reform.   


%% file: 1_background.tex
\vspace{-2mm}
\section{Background}



\subsection{Anatomy of a learning process}
\label{sec:anatomy}
An idealized learning process begins with the formulation of \textbf{goals} (including scientific goals such as understanding what factors influence a particular human behavior, engineering goals such as constructing a better model for machine translation, or policy goals such as estimating effectiveness among different types of patients) and \textbf{hypotheses}. These are not necessarily statistical ``hypotheses''; rather, a hypothesis could be that a certain thinking pattern increases the chances of a behavior, or that a certain technical innovation will lead to a better translation system, or that a treatment will be more effective among men than women.  Goals and hypotheses lead to steps of
\textbf{data collection and preparation}. Researchers specify an observational process to collect information about the latent phenomena of interest from the environment. An observational probe is used to induce explicit observations thought to be sensitive to the target phenomena. For example, psychologists design human subjects experiments using interventions thought to interact with the target phenomena. ML researchers often make use of datasets generated from human produced media and signals of behavior, in the form of digital traces. 

An observational process becomes a model by making assumptions about what the observed data represent, namely realizations of random variables with regular variation. The observational model is defined by a \textbf{model representation}, i.e., the model class or functional form that specifies a space of data generating processes (DGPs, i.e., fitted functions) that might have produced the data. This might be a multiple linear regression functional form in social psychology, or a more DNN architecture in ML (with a specific configuration of network parameters like arrangement into convolutions, activation functions, etc.). Because quantifying and searching \textit{all} DGPs implied by probability distributions over the observation space tends to be prohibitively complex, learning pipelines often consider a subset or ``small world'' of model configurations~\cite{bernstein2021}, called the hypothesis space of the learner in ML. \textbf{Model selection} or model-based inference describes how a best fit model that is most consistent with the data is determined. This involves defining an objective or loss function measuring the difference between the ground-truth observed outcome for an input and the predicted outcome of a parameterized model configuration (e.g., squared error), as well as an optimization method for searching the space of model configurations to find the fitted function that minimizes loss (e.g., gradient descent, adaptive optimization algorithms, analytical solutions like maximum likelihood estimation (MLE), etc.). 

An \textbf{evaluation} may follow to validate the usefulness of what is learned relative to alternative model fits or learning pipelines. Evaluation metrics such as explained variance or log loss can be used to summarize overall usefulness of a fitted function. Evaluation metrics may sometimes be implicit, such as when the usefulness of a fitted model is evaluated relative to one's hypotheses about the data generating process. 
The learning process culminates in \textbf{communication} of claims in research papers. By ``errors in learning,'' we refer to issues that arise in this larger process in which a researcher specifies and ``solves'' a learning problem. 

\vspace{-6mm}\subsection{Goals of learning in social psychology versus machine learning}
\label{sec:goals}

\textbf{Social psychology.} A primary goal in empirical psychology is to describe the causal underpinnings of human behavior~\cite{meehl1967theory,shmueli2010explain,yarkoni2017choosing}. Researchers identify hypotheses representing predictions about variables that constitute observed data. Often these constitute ``weak theories”~\cite{meehl1990summaries}, predicting a directional difference or association between variables but not the size of the effect. They design observational processes to gather data for testing hypotheses, typically controlled human-subjects experiments that record the thoughts, emotions, or behavior of subjects, under different conditions thought to interact with the latent phenomena of interest. The approximating functions that researchers learn from these observations (often low dimensional linear regressions) are thought to capture key structure in the latent psychological phenomena. Claims about cause and effect hinge on interpreting the parameter values of the fitted function in light of hypotheses and their sampling variation within a statistical testing framework. A function is commonly deemed worthy of interest if it's p-value is below a false-positive rate defined in the Neyman-Pearson framework, typically $\alpha=0.05$. Direct claims take the form of statements about novel, statistically significant causal attributions, and have been called ``stylized facts''~\cite{gelman2018failure,hirschman2016stylized} implied by authors to be generally true about human behavior. For example, thinking about old age induces old-like behavior~\cite{bargh1996automaticity}.

\textbf{Machine learning.} A primary goal in supervised ML research is to facilitate the learning of functions which achieve high predictive accuracy in tasks like classification. Researchers hypothesize procedures or abstractions that may improve the state-of-the-art (SOTA) in subareas (e.g., natural language processing (NLP), vision), which is captured by benchmarks: abstractly defined tasks (e.g., image classification, machine translation) instantiated with learning problems consisting of datasets (input, output pairs) and an evaluation metric to be used as a scoring function (e.g., accuracy)~\cite{liao2021we}. Standard methods like using a train-test split and cross validation are designed to ensure good predictive performance of a fitted model on unseen data. 
Claims in empirical research papers typically report performance of a new learner (i.e., model) on benchmarks, compared to baselines representing the prior SOTA. Formal proofs of the statistical properties of new methods are also common.

%% file: 2_1_data.tex
\section{Threats to learning in social psychology and machine learning}

We describe threats to valid learning according to whether they involve data selection and preparation, model development (including choosing a representation and a model selection and evaluation approach), and communication of results in a research paper. 

\vspace{-2mm}
\subsection{Data collection and preparation}
\label{sec:data}
\textbf{Social psychology.} {\em High measurement error relative to signal, unacknowledged flexibility in defining data inputs, underspecified or non-representative subject samples, and underspecification of stimuli generation, and other ``design freedoms'' can threaten the validity of conclusions drawn in empirical psychology research.}

The design of many psychology experiments implies that researchers do not grasp the implications of using small samples and noisy measurements to draw inferences about effects that are a priori likely to be small. For example, a thought-to-be pervasive belief is that if an experiment registers a ``statistically significant'' effect on a small sample, then that effect will necessarily remain significant with a larger sample~\cite{simmons2011false,button2013power}. In reality, with a lower powered study, not only is there a lower probability of finding a true effect of a given size, but there is a lower probability that an observed effect which passes a significance threshold actually reflects a true effect that will appear under replication~\cite{button2013power}. 
Under low power, estimates of observed effects will tend to reflect sampling error that derives from the limited size of the sample relative to a target population, and forms of measurement error~\cite{loken2017measurement}, such as random variation due to noise in taking measurements that produces a difference between observed and true values. Studies are ``dead on arrival'' when standard error due to measurement and sampling variation is large relative to any plausible effect size~\cite{gelman2009beauty}.

Inherent flexibility in how a researcher specifies an analysis is a different type of threat. A ``researcher degrees of freedom'' or ``garden of forking paths'' metaphor~\cite{simmons2011false,gelman2014statistical} suggests that human tendencies toward self-serving interpretations of ambiguous evidence (e.g.,~\cite{babcock1997explaining,dawson2002motivated} as cited in~\cite{simmons2011false}), make researchers likely to draw conclusions that verify their hypotheses. 
Given an outcome of interest (e.g., self-reported political preference), an analyst may bias results toward a preferred conclusion by selecting data transformations and outlier removal processes, or choosing between different predictor variables or ways of operationalizing the outcome variable, conditional on seeing the results of these choices, without necessarily recognizing they are doing anything improper.  
More broadly, when a researcher can tweak the design of experiment conditions with feedback through pilot experiments via the design of stimuli, instructions, and elicitation instruments, they may gravitate toward designs that exaggerate effects in some conditions, resulting in a form of procedural overfitting.  

Scholars have pointed to study results not being reproducible because they use non-representative samples of a target population, such as convenience samples of university students from western educated industrialized rich democratic (WEIRD) countries~\cite{henrich2010weirdest}. As researchers have become more accustomed to the importance of statistical power and representative samples, online recruitment of participants in social psychology~\cite{sassenberg2019research} has increased. However, it is unclear that sample homogeneity is addressed by online samples~\cite{chandler2014nonnaivete} and this trend has led to greater use of self-reported measures~\cite{sassenberg2019research} that contribute additional noise. 
More generally, failure to recognize the implications of non-random sampling can lead to a ``big data paradox'' of overconfidence as sample size increases~\cite{meng2018statistical}. 
Another fundamental but often overlooked issue concerns how psychologists often leave the target population of their inferences unspecified~\cite{gigerenzer2015surrogate}, making it ambiguous what is being learned at all. 

\textbf{Machine learning.} {\em Standardization of benchmarks and the prohibitive cost of amassing large datasets means that researchers often rely on existing datasets~\cite{halevy2009unreasonable,sun2017revisiting}, typically obtained through crowdsourced annotation and web-scale data (e.g.,~\cite{deng2009imagenet,krizhevsky2009learning}). Similar to psychology, factors like choosing how to transform data after seeing results, the use of non-representative samples, and underspecification of the population captured in data threaten the validity of claims. More frequently discussed issues include the differential effects of non-random measurement error on real world outcomes when a model is deployed and the way that a ``good'' predictive model can perpetuate forms of historical bias like stereotypes.}

Recent work in ML points to analogous concerns to psychology in recent acknowledgement of flexibility in data transformation, such as in filtering data in ways that simplify a prediction problem (e.g., removing translation artifacts in machine translation to improve prediction accuracy~\cite{liaoforward} as cited in \cite{liao2021we}). 

Non-representative samples are also a concern, including violations of the assumption that the development distribution from which the training and test data are presumed to be randomly drawn is the same as the deployment distribution from which samples will be drawn in real-world applications~\cite{suresh2021framework,paullada2021data,barocas2016big}. 
`Representation bias''~\cite{suresh2021framework} involves development data that underrepresent some parts of the input space of an ML algorithm, leading to higher error rates for less-represented instances in the input space (e.g.,~\cite{buolamwini2018gender,park2018reducing,zhao2017men}). Suresh and Guttag~\cite{suresh2021framework} define this bias as a positive value for a measure of divergence between the probability distribution over the input space and the true distribution, noting that it can occur simply as a result of random sampling from a distribution where some groups are in the minority. 
Others describe how error in the (often unreported~\cite{gebru2021datasheets}) labeling process used to construct ground truth can lead to overfitting~\cite{bowman2015large,northcutt2021pervasive}, as well as how data preparation steps lose information whenever majority-rule is used to construct a ground truth without preserving information about label distributions (e.g., describing variance across annotators) ~\cite{davani2022dealing,gordon2021disagreement}.

However, criticism of data practices in ML often focuses on systematic measurement error (i.e., bias) in collected data that threatens construct validity: whether the measurement is actually capturing the intended concept. 
``Measurement bias''~\cite{suresh2021framework} has been used to refer to differential measurement error~\cite{vanderweele2012results}, where a measurement proxy is generated differently across groups due to differing granularity or quality of data across groups, or reduction of complex target category (e.g., academic success) to a small number of proxies that favor certain groups over others (e.g.,~\cite{kleinberg2018algorithmic} as cited in~\cite{suresh2021framework}). 
Jacobs and Wallach~\cite{jacobs2021measurement} attribute many misleading claims in the fairness literature in ML to unacknowledged mismatches between unobservable theoretical constructs in ML applications (e.g., risk of recidivism, patient benefit) and the measurement proxies that researchers often tend to assume capture them, and suggest the use of latent variable models to formally specify assumptions. 


A novel concern about measurement bias in ML relative to psychology occurs when biased input data are used to train a model and contribute to undesirable social norms. Data may record historical biases~\cite{suresh2021framework} (e.g., training a model to recognize successful applicants on data where women were admitted less due to bias). ``Harms of representation''~\cite{crawford2017,abbasi2019fairness} refers to how model predictions can reinforce potentially harmful stereotypes when trained on data exhibiting bias. For example, returning pictures of only white males on a Google search for CEO reinforces notions that other groups are not as appropriate for CEO positions~\cite{kay2015unequal}. The fact that ML is often intended for prescriptive use in the world, rather than descriptive use as in psychology research helps explain the prevalence of these concerns and the emphasis on systematic measurement error. 

Finally, data concerns in ML increasingly refer to forms of underspecification of population details and underacknowledgment of the constructed nature of data, instead taking data as given~\cite{bender2021dangers,gebru2021datasheets,scheuerman2021datasets,hutchinson2021towards}. These concerns also imply that real-world harms may result from practices that extract away the subjective judgments, biases, and contingent contexts involved in dataset production~\cite{paullada2021data}. 

%% file: 2_2_model.tex
\begin{table*}[!h]
  \centering 
  \begin{adjustbox}{width=\textwidth,center}
  \begin{tabular}{P{2.3cm}P{7cm}P{8.5cm}}
     & \textbf{Social Psychology} & \textbf{Machine Learning} \\
    \midrule
    \textbf{Data selection and preparation} & 
    \begin{minipage}[t]{\linewidth}
    \begin{itemize}[leftmargin=*,nosep,after=\strut]
        \item[$\circ$] \sethlcolor{mypink1}\hl{High measurement error relative to the size of effects being studied~\mbox{\cite{loken2017measurement,baumeister2007psychology,devezer2020case}}}\tikzmarkhf{l12}
        \item[$\circ$]\sethlcolor{myblue1}\hl{Data transformations decided contingent on (NHST) results~\mbox{\cite{simmons2011false,gelman2014statistical}}}\tikzmarkhf{l13}
        \item[$\circ$]\sethlcolor{mygreen1}\hl{Non-representative~\mbox{\cite{henrich2010weirdest,meng2018statistical}} or underdefined samples~\mbox{\cite{gigerenzer2015surrogate}}; insufficient stimuli sampling~\mbox{\cite{wells1999stimulus,yarkoni2022generalizability,gigerenzer2022we}}}\tikzmarkhf{l14}
        \item[$\circ$]Small samples and noisy measurements (low power) leading to biased estimates~\cite{button2013power}\tikzmarkhf{l11}
    \end{itemize}
    \end{minipage} & 
    \begin{minipage}[t]{\linewidth}
    \begin{itemize}[leftmargin=*,nosep,after=\strut]
        \item[$\circ$]\tikzmark{r11}\sethlcolor{mypink1}\hl{Differential measurement error (e.g., across social groups)~\mbox{\cite{suresh2021framework,buolamwini2018gender,park2018reducing,zhao2017men}} which is not modeled~\mbox{\cite{jacobs2021measurement,kleinberg2018algorithmic}}}
        \item[$\circ$]\tikzmark{r14}\sethlcolor{mypink1}\hl{Label errors~\mbox{\cite{bowman2015large,northcutt2021pervasive}} and disagreement~\mbox{\cite{davani2022dealing,gordon2021disagreement}}}
        \item[$\circ$]\tikzmark{r12}\sethlcolor{myblue1}\hl{Data transformations decided contingent on performance comparisons~\mbox{\cite{liaoforward,bone2015applying}}}
        \item[$\circ$]\tikzmark{r13}\sethlcolor{mygreen1}\hl{Underrepresentation of portions of input space in training data~\mbox{\cite{suresh2021framework,paullada2021data,barocas2016big}}}
        \item[$\circ$]Input data too huge to understand~\cite{bender2021dangers,paullada2021data}
    \end{itemize}
    \end{minipage} \\
    \midrule
    \textbf{Model representation} & 
    \begin{minipage}[t]{\linewidth}
    \begin{itemize}[leftmargin=*,nosep,after=\strut]
        \item[$\circ$] \sethlcolor{mypink1}\hl{Overreliance on models and designs with good asymptotic guarantees~\mbox{\cite{normand2016less}}}\tikzmark{l22} 
        \item[$\circ$]\sethlcolor{myblue1}\hl{No explicit representation of prior/domain knowledge~\mbox{\cite{prior2017,gelman2012ethics}}}
        \item[$\circ$]\sethlcolor{myblue1}\hl{Inappropriate expectations~\mbox{\cite{clark1973language,gelman2015connection,yarkoni2022generalizability}}} in light of crud factor~\mbox{\cite{meehl1990summaries,orben2020crud}}; belief in many nudging factors with large consistent effects on outcome~\mbox{\cite{tosh2021piranha}}
        \item[$\circ$]\sethlcolor{myblue1}\hl{Unacknowledged multiplicity of solutions~\mbox{\cite{yarkoni2017choosing}}}\tikzmark{l23}
        \item[$\circ$]Structural misspecification~\cite{vowels2021misspecification,lynch1982external}\tikzmark{l21}
    \end{itemize}
    \end{minipage} & 
    \begin{minipage}[t]{\linewidth}
    \begin{itemize}[leftmargin=*,nosep,after=\strut]
        \item[$\circ$]\tikzmark{r22}\sethlcolor{mypink1}\hl{Overreliance on asymptotic (worst-case) guarantees~\mbox{\cite{domingos2012few}}}
        \item[$\circ$]\tikzmark{r23}\sethlcolor{myblue1}\hl{Underspecification of desired inductive biases~\mbox{\cite{d2020underspecification,ilyas2019adversarial}}; failure to prevent shortcut learning~\mbox{\cite{geirhos2020shortcut}}}
        \item[$\circ$]\tikzmark{r21}Inappropriate i.i.d. assumption in light of real-world nonstationarity~\cite{widmer1996learning,bickel2009discriminative,storkey2009training}
        \item[$\circ$]\tikzmark{r24}Reliance on fine-tuning/foundation models for which hyperparameter tuning is opaque~\cite{dodge2020fine,wortsman2021robust}
        \item[$\circ$]\tikzmark{r25}Convergence in architectures around large models~\cite{bommasani2021opportunities,bender2021dangers,strubell2020energy}
    \end{itemize}
   \end{minipage} \\
    \midrule
    \textbf{Model selection and evaluation} & 
    \begin{minipage}[t]{\linewidth}
    \begin{itemize}[leftmargin=*,nosep,after=\strut]
        \item[$\circ$]\sethlcolor{mypink1}\hl{Implicit optimization for statistical significance~\mbox{\cite{goodman1993p,lee2014bayesian,hubbard2003confusion,gelman2014beyond,gelman2014ethics,gelman2014statistical}}}\tikzmark{l33}
        \item[~]
        \item[$\circ$]Inference as black box~\cite{gigerenzer2015surrogate,lee2014bayesian,wasserman2004bayesian}; Not motivating choice of estimator or optimization for particular inference goal~\cite{berger1988likelihood,wagenmakers2018bayesian}\tikzmark{l31}
        \item[$\circ$]Misunderstanding/misusing ideas of statistical significance~\cite{hubbard2003confusion,hubbard2003p,wagenmakers2007practical,gelman2013commentary,greenland2019valid}\tikzmark{l34}
        \item[$\circ$]Multiple comparisons problem~\cite{gelman2013garden}\tikzmark{l35}
    \end{itemize}
    \end{minipage} & 
    \begin{minipage}[t]{\linewidth}
    \begin{itemize}[leftmargin=*,nosep,after=\strut]
        \item[$\circ$]\tikzmark{r36}\sethlcolor{mypink1}\hl{Implicit optimization to beat SOTA~\mbox{\cite{hofman2017prediction,sculley2018winner}}}
        \item[$\circ$]\tikzmark{r31}Knowledge of how OOD test sets are constructed used to choose representation/method~\cite{teney2020value}
        \item[$\circ$]\tikzmark{r31}Overlooked sensitivity of optimizer performance to hyperparameters~\cite{bouthillier2020survey,choi2019empirical,schmidt2021descending,gilmer2021loss,picard2021torch}; computational budget~\cite{teja2019optimizer}
        \item[$\circ$]\tikzmark{r32}Presence of implementation variation~\cite{liao2021we} and tricks~\cite{andrychowicz2020matters,henderson2018deep}
             \item[$\circ$]\tikzmark{r35}Misuse of cross validation~\cite{cawley2010over,hastie2009elements,bergmeir2012use,teney2020value,hosseini2020tried}
              \item[$\circ$]\tikzmark{r34}Optimism of cross validation~\cite{efron2004estimation,malik2020hierarchy}
        \item[$\circ$]\tikzmark{r33}Loss metric misalignment~\cite{huang2019addressing}
     
        \item[$\circ$]\tikzmark{r37}Not comparing to simpler baselines~\cite{sculley2018winner,dacrema2021troubling} or priors~\cite{goyal2017making}
    \end{itemize}
    \end{minipage} \\
    \midrule
    \textbf{Communication of claims} & 
    \begin{minipage}[t]{\linewidth}
    \begin{itemize}[leftmargin=*,nosep,after=\strut]
        \item[$\circ$]\sethlcolor{mypink1}\hl{Unwarranted speculation about what evidence a p value provides~\mbox{\cite{szucs2017null}}}\tikzmark{l42}
        \item[$\circ$]\sethlcolor{myblue1}\hl{Overgenerationalization (i.e., beyond studied population)~\mbox{\cite{dejesus2019generic,gutierrez2003cultural,rogoff2003cultural,yarkoni2022generalizability,simons2017constraints}}}\tikzmark{l43}
        \item[$\circ$]\sethlcolor{mygreen1}\hl{Unavailable data and code~\mbox{\cite{honesty2017,houtkoop2018data,munafo2017manifesto}}}\tikzmark{l45}
        \item[$\circ$]Not acknowledging having explored multiple analyses conditioned on data~\cite{simmons2011false,gelman2013garden}\tikzmark{l41}
        \item[$\circ$]Inaccurate descriptions of what p values mean~\cite{besanccon2019continued,altman1995statistics,gelman2006difference,szucs2017null}\tikzmark{l44} 
    \end{itemize}
    \end{minipage} & 
    \begin{minipage}[t]{\linewidth}
    \begin{itemize}[leftmargin=*,nosep,after=\strut]
        \item[$\circ$]\tikzmark{r43}\sethlcolor{mypink1}\hl{Unwarranted speculation about causes~\mbox{\cite{liao2021we,lipton2019research,kaushik2018much}}}
        \item[$\circ$]\tikzmark{r44}\sethlcolor{myblue1}\hl{Implying equivalence of learning problems and human performance on a task~\mbox{\cite{kaushik2018much,liao2021we,lipton2019research}}}
        \item[$\circ$]\tikzmark{r45}\sethlcolor{mygreen1}\hl{Lack of dataset documentation~\mbox{\cite{bender2021dangers,gebru2021datasheets,paullada2021data}}}
        \item[$\circ$]\tikzmark{r46}\sethlcolor{mygreen1}\hl{Inaccessible data, code, computational resources~\mbox{\cite{gorman2019we,sandve2013ten,stodden2014provisioning}}}
        \item[$\circ$]\tikzmark{r41}Not reporting implementation conditions/sources of variance~\cite{sculley2018winner,lipton2019research}
        \item[$\circ$]\tikzmark{r42}Underpowered performance comparisons~\cite{agarwal2021deep,bouthillier2020survey}; ignoring sampling error~\cite{agarwal2021deep,liao2021we,raschka2018model}
    \end{itemize}
    \end{minipage} \\
    \bottomrule

  \end{tabular}
  \end{adjustbox}
    \caption{\textbf{Overview of learning concerns}, roughly ordered to emphasize similarities across social psychology and ML.}
    \label{tab:results}
\end{table*}
\vspace{-8mm}
\subsection{Model representation}
\label{sec:model}
Learning from data requires selecting a model representation, a formal representation that defines what functions can be learned. 

\textbf{Social psychology.} {\em Researchers commonly overlook the importance that the small world of model configurations they explore captures or well approximates the true DGP for valid inference, hold unrealistic views about the separability of large effects in the world, and tend to incorporate prior knowledge into modeling informally rather than explicitly.} 

When modeling a latent psychological phenomenon, often via simple measures of correlation and linear parametric models~\cite{blanca2018current,bolger2019causal}, researchers implicitly assume that there is a true DGP that exactly captures how the target arises as a function of other factors thought to influence it. Once an observational model 
is defined, inference is confined to the mathematical narratives represented by these functions~\cite{prior2017}. However, the validity of claims made about causal effects by following this process depend upon judicious choices about how to represent structure in the true DGP in the constrained small world model space, which psychology researchers often overlook~\cite{vowels2021misspecification}. 

A first complication arises from the fact that inference is more straightforward when the true DGP is included in the small world of configurations under consideration~\cite{bernardo1994}. However, the sorts of human behaviors psychologists tend to target are thought to be conceptualizable but too complicated to specify explicitly, or not even conceptualizable~\cite{vowels2021misspecification}. Under these conditions, the validity of conventional interpretations of fitted models depends on the observational model faithfully approximating the true DGP~\cite{prior2017}. 

However, this is not the case when a model is structurally misspecified, meaning the fitted models do not adequately capture the true causal structure and/or the functional form of the relationships between variables in the true DGP~\cite{vowels2021misspecification}. For example, if the DGP in a psychology study can be described as a weighted sum of the set of input variables that are represented in the chosen functional form, and all of these predictors are exogenous (i.e., completely independent), then parameters estimated using ordinary least squares can be interpreted according to convention as information about the target phenomena (e.g., comparing two items that differ by one unit in predictor $x$ while being the same in all other predictors will differ in $y$ by $\theta$, on average). However, when the true DGP is more complex than the functional form, the choice of which potential confounding variables one measures and includes in the regression equation becomes important. Not including variables that influence a regressor and the outcome~\cite{angrist2008mostly} or including variables that could in principle be affected by experimental manipulations (and hence represent outcome variables themselves~\cite{coyle2020targeting}) cause the conventional interpretation of the fitted parameter values not to hold. However, researchers seldom acknowledge these limitations.
 

Researchers often choose designs based on a preference for simpler models. Perhaps the most common example is preferring between-subject designs 
based on their asymptotic properties: as the size of the (random) sample increases toward the population size, a between-subjects design provides a simpler procedure for estimating average treatment effects relative to a within-subjects design, which requires estimating carryover effects between treatments experienced by the same individual~\cite{normand2016less}. However, high variation between people can lead to poor estimates of average treatment effects if the treatment interacts with background variables associated with differences in individuals and contexts~\cite{lynch1982external,normand2016less}. 

More generally, psychology researchers have been criticized for estimating effects as if they are constant rather than assuming they will vary across people or contexts~\cite{gelman2015connection}. 
This can manifest, for example, as model specifications that ignore the importance of modeling variation in stimuli and other experimental conditions as well as subjects~\cite{yarkoni2022generalizability} (e.g., a ``fixed effect fallacy''~\cite{clark1973language}).

Tendencies to overlook important sources of variation in modeling are implied by
Meehl’s conception of the ``crud factor''~\cite{meehl1990summaries,orben2020crud}, which emphasizes how causal attribution using constrained model spaces to approximate a highly complex true DGP is fundamentally challenged by the prevalence of ``real and replicable correlations'' reflecting ``true, but complex, multivariate and non-theorized causal relationships'' between all variables~\cite{orben2020crud}. 
Problems arise when researchers overlook model misspecification due to conventional but questionable beliefs about reality. For example, a tendency toward reporting model fits suggesting that novel yet seemingly trivial ``nudging'' factors (e.g., whether or not someone is menstruating~\cite{durante2013fluctuating} or whether there was a recent shark attack~\cite{achen2012blind}) have large and consistent effects on the same outcomes (e.g., voting behavior) overlooks the fact that if 
such effects were large, we should expect them to interact in complex ways. Hence, we should expect it to be very difficult to observe stable and replicable effects~\cite{tosh2021piranha}. 

In this way, choices of model representation (i.e., low dimensional linear regressions) are not fully consistent with prior knowledge. Conventional approaches to estimating an effect of interest are also memoryless in the sense that even when prior estimates of an effect of interest are available, e.g., from past experiments, they are generally not incorporated in the model representation. 
Combined with incentives to publish surprising results~\cite{fanelli2010positive,scargle1999publication} and the inflated probability of observed effects to be overestimates in small sample size studies (Section~\ref{sec:data}), this can result in published effects that seem suspiciously big in light of prior domain knowledge. 



\textbf{Machine learning.} {\em 
In theory, optimizing for predictive accuracy does not require well-approximating a true DGP. However, researchers' commonly assume that unseen data are drawn from the same distribution as training data and use asymptotics to motivate model choice, leading to unrealistic beliefs about the predictability of real world processes. Threats also arise from failures to explicitly represent a priori human expectations about what predictors are valid for a task, and a convergence on hard-to-analyze models that combine pre-trained representations with domain-specific data. } 

The biggest point of contrast between representations in supervised learning in ML and social psychology is that the former traditionally do not assume that the learning process is ``realizable''~\cite{russell2003artificial} in the sense that the true DGP is in the set of learnable functions (or hypothesis space), nor even that the fitted function approximates the structure of the true DGP. Instead, the goal of learning can be formulated as identifying a function with error that can be guaranteed to fall within some bound of the best possible predictor over possible samples~\cite{valiant1984theory}. Choosing a representation (i.e., hypothesis space) in theory means reasoning about the inductive biases (i.e., properties of the predictors) it will return in light of prior knowledge, but in practice theoretical guarantees (e.g., worst-case bounds) on convergence or generalization ability have driven representation choices. This can lead, as in psychology, to use of models in cases where asymptotic assumptions don't apply~\cite{domingos2012few}. Or, as in the case of much recent DL research, a more empirical, performance-driven approach uses achievable performance as the primary driver of model choice~\cite{recht2018cifar}.
One of the most commonly cited deficiencies attributed to model representations in applied ML involves assuming a static relationship between the predictor variables and the outcome~\cite{vapnik1998statistical}, which supports conventions like shuffling input data to create training and test sets~\cite{arjovsky2019invariant}. This assumption makes models vulnerable to concept drift~\cite{widmer1996learning} (a.k.a. covariate shift~\cite{bickel2009discriminative} or distribution or dataset shift~\cite{storkey2009training}), where predictions are inaccurate post-hoc due to non-stationarity in the real-world relationship between the inputs and outputs due to temporal changes (e.g.,~\cite{luu2021time}), behavioral reactions (e.g.,~\cite{perdomo2020performative}), or other unforeseen dynamics~\cite{quinonero2008dataset}. 
Under conventional ``distribution unawareness,'' it also becomes difficult to distinguish when unexpected errors arise from distribution shift versus inefficiencies in the learning pipeline~\cite{berend2020cats}.
Distribution shift can lead to poorly calibrated estimates of the uncertainty of model performance~\cite{ovadia2019can}, similar to how choosing estimators by convention rather than guided by one's inference goal (see Section~\ref{sec:eval}) biases uncertainty estimates for effects observed in psych experiments.

Distribution shift motivates greater focus on how different models fare at out-of-distribution (OOD) error and their robustness to adversarial manipulation, i.e., small changes to an input in feature space that dramatically change the predicted output (e.g.,~\cite{szegedy2013intriguing,beery2018recognition,rosenfeld2018elephant,carlini2017towards,teney2020value}).
Recent results related to adversarial nonrobustness~\cite{ilyas2019adversarial}, underspecification~\cite{d2020underspecification}, shortcut learning~\cite{geirhos2020shortcut}, simplicity bias~\cite{shah2020pitfalls}, and competency problems~\cite{gardner2021competency} suggest that beliefs about the true DGP in predictive modeling as in ML are not necessarily as distinct from explanatory, attribution-oriented modeling as past comparative accounts (e.g., ~\cite{breiman2001statistical}) imply. 

For example, one understanding of concept drift that we can relate to the so-called crud factor in psychology is that the concept of interest (or target task) in an ML pipeline for discriminative learning often depends on a complex combination of features that are not explicitly represented in the model. Geirhos et al.~\cite{geirhos2020shortcut} use ``shortcut learning'' to refer to a tendency for ML models to learn simple decision rules (e.g.,~\cite{jo2017measuring,mccoy2019right,arpit2017closer}) that perform well on standard benchmarks. While these features represent ``real'' correlations, the problem is that singular predictive features mined in training data often do not perform as well in more challenging testing situations, where a human might naturally expect successful performance to require combinations of features (e.g., derived from several different object attributes in object recognition).
Shortcut learning and related vulnerabilities to adversarial manipulation imply not a failure in learning from a modeling standpoint, nor even a failure of a fitted function to generalize~\cite{geirhos2020shortcut}, but a mismatch between a human's conception of critical, stable properties that predict under the true DGP and those that drive the predictions of the fitted model~\cite{d2020underspecification,geirhos2020shortcut,ilyas2019adversarial}. 

A related theory representing a symptom of predictive multiplicity is underspecification~\cite{d2020underspecification}: specifically, a failure to represent in the learning pipeline which inductive biases are more desirable to constrain learning. Underspecification occurs when predictors with equivalent performance on i.i.d. data from the same distribution as training degrade non-uniformly in performance when probed along practically relevant dimensions~\cite{d2020underspecification}. Underspecification is distinct from forms of distribution shift that may give rise to shortcut learning, such as the presence of spurious features in the training data that are not associated with the label in other settings. Instead, it captures how a single learning problem specification can support many near-optimal solutions but which might have different properties along some human relevant dimensions like fairness or interpretability~\cite{ross2017neural}.  

A common approach to overcoming poor generalization of a model is to combine multiple representations. 
Representation learn\-ing---automated, untrained learning of input representations (i.e., generic priors) on huge datasets that capture structure in domains like language or vision---reduces the difficulty of achieving high accuracy in domains where labeled data is costly~\cite{bengio2013representation}. ``Fine-tuning'' pretrained ``foundation'' models~\cite{bommasani2021opportunities} for domain-specific applications has become standard practice based on the performance that can be achieved over conventional domain-specific learning pipelines~\cite{wortsman2021robust,sun2017revisiting,gururangan2020don}. Though training on highly diverse input data tends to provide foundation models with inductive biases that improve extrapolation, a challenge is that fine-tuning performance can be highly sensitive to how poorly-understood parameters are set, making results hard to replicate~\cite{dodge2020fine}. 
For example, 
the robustness of a fine-tuned model has been found to vary considerably under small changes to hyperparameters~\cite{wortsman2021robust}.
Related is a concern that the convergence in deep learning research around large DNN model architectures with minimal task-specific parameters~\cite{bommasani2021opportunities} 
doubles down on an approach that imposes unreasonable environmental~\cite{bender2021dangers,strubell2020energy} and research opportunity costs~\cite{bender2021dangers}.

More generally, understanding the implications of model selection is complex for DNNs, where classical theory falls short of explaining the generalization performance (e.g.,~\cite{zhang2021understanding,jiang2019fantastic,belkin2018overfitting,belkin2019reconciling,dar2021farewell}). This has motivated lines of theoretical work that explore different explanations of phenomena like ``double descent''~\cite{belkin2019reconciling}, where the generalization performance of a deep model continues to improve even after it has achieved zero loss on (or perfectly interpolated) the training data. For example, some analyze the properties of overparameterized linear regressions~\cite{dar2021farewell}. 

%% file: 2_3_objective_etc.tex
\vspace{-3mm}
\subsection{Model selection and evaluation}
\label{sec:eval}
Model-based inference involves explicit and implicit choices of objective function, optimization approach, and evaluation metric. 

\textbf{Social psychology.} 
{\em Claims made in social psychology research are threatened when researchers treat conventional approaches to model-based inference as a black box for consuming data and outputting inferences~\cite{bakan1966,gigerenzer2015surrogate,lee2014bayesian}, and by researchers' implicit use of statistical significance alone as a criterion for deciding what to report.}

It is relatively rare for psychology research contributions to include explicit motivation for the estimators and loss functions used in modeling. Such ``inference by convention'' can produce misleading claims without outright cheating or motivated reasoning similar to how blindly preferring between-subjects designs can. For example, conventional use of maximum likelihood estimators based on their consistency~\cite{wasserman2004bayesian} may lead researchers to overlook critical assumptions required for these estimators to be well-calibrated (i.e., have sampling distributions which are asymptotically normal). Analytical approaches to optimization bring convenience, 
but commonly-used approaches to model fitting and selection tend to be based on pre-experimental guarantees (i.e., before data are collected), which cannot guarantee that they will be appropriate (e.g., well-calibrated) on a particular dataset~\cite{berger1988likelihood,wagenmakers2018bayesian}. 


A different source of misleading claims is the use of statistical significance as a coarse objective function. 
Implicit optimization for significance, in which researchers are essentially searching through a garden of forking paths for specifications that achieve significance as a sort of quasi-optimization approach~\cite{gelman2014statistical}, means that conventional interpretations of fitted models and statistical tests on parameter estimates will not hold. For example, the multiple comparisons problem, in which researchers neglect to control for data-dependent selection in what they report, alters the statistical properties of estimates and tests~\cite{gelman2013garden}. 
At the highest level, bias affects the published record when researchers decide whether to report results based on the significance levels~\cite{scargle1999publication,gelman2014beyond}. 

Using ``statistical significance'' as an implicit objective does not line up with scientific goals (e.g.,~\cite{hubbard2003confusion,hubbard2003p,wagenmakers2007practical,gelman2013commentary,greenland2019valid}). 
For example, the use of $p$-values and statistical significance in psychology research is described as fundamentally confused in that rejection of straw-man null hypotheses is inappropriately taken as evidence in favor of researchers' preferred alternatives~\cite{goodman1993p,lee2014bayesian,hubbard2003confusion}. In other words, hypothesis testing can sometimes be used as a sort of ``truth mill'' in psychology~\cite{gelman2014beyond,gelman2014ethics}. 

Related problems include not acknowledging that as a random variable, $p$ can vary considerably even under idealized replication~\cite{goodman1993p,gelman2006difference,senn2001two,murdoch2008p,boos2011p}, such that the difference between significant and not significant is not itself significant. Researchers also overlook the fact that for $p$ to be a valid estimate of the probability of observing an effect as large or larger than that seen, all assumptions about the test and observational process must hold~\cite{amrhein2019inferential,greenland2019aid,rafi2020semantic}.

\textbf{Machine learning.} {\em 
Claims are often made in ML research without acknowledging that they depend critically on choices of hyperparameters, initial conditions, and other configuration details that directly influence performance in non-convex optimization.  
Researchers also may exploit flexibility in designing performance comparisons in order to 
achieve superior performance for their contributed approach relative to alternatives~\cite{hofman2017prediction,sculley2018winner}. 
}


In contrast to loss functions in simple regression models, ML models tend to have high dimensional non-convex loss. While this does not necessarily prevent generalization~\cite{choromanska2015loss}
it makes solutions like saddle points, which can give the illusion of a satisfactory local minimum, of greater concern~\cite{dauphin2014identifying,saxe2013exact}. 
Optimizers---algorithms that prescribe how to update parameter values like weights during inference to reduce the value of the objective on the training data---are critical to the accuracy gains seen in recent years.
However, to make non-convex optimization tractable requires setting various opaque hyperparameters and initial conditions that influence how the loss landscape is traversed~\cite{choi2019empirical,schmidt2021descending,gilmer2021loss}. 


For an optimization approach like stochastic gradient descent (SGD), hyperparameters like the learning rate affect how quickly it learns the local optima of a function: too high a rate means the function cannot converge, too low and it may require too long~\cite{bouthillier2020survey}. 
Adaptive optimizers (e.g., Adagrad, Adam) allow hyperparameters like learning rate to vary for each training parameter, inducing a new dynamical system with each run and complicating attempts to explain 
what parts of a pipeline improved performance. 

Hyperparameter tuning is also a computationally expensive task~\cite{teja2019optimizer}, inducing uncertainty about how a solution might differ under a larger computational budget or different parameter settings. 
Some recent work finds that given a fixed computational budget, choosing the best optimizer for a task with the default parameters performs about as well as choosing any widely-used optimizer and tuning its hyperparameters, questioning claims of state-of-the-art performance of newly introduced optimizers across tasks~\cite{schmidt2021descending}. Similarly, sufficient hyperparameter optimization can mostly eliminate claimed performance differences in generative adversarial networks (GANs)~\cite{lucic2018gans}, and better hyperparameter tuning on baseline implementations can eliminate evidence of performance advantages of new learning methods~\cite{henderson2018deep,melis2017state}. 
Liao et al.~\cite{liao2021we} use the broader term ``implementation variation'' to refer to how variations in how inference techniques are implemented---including use of specific software frameworks and libraries, metric scores, and implementation ``tricks'' ~\cite{andrychowicz2020matters,henderson2018deep}---can affect their performance in evaluations. 
A related concern in subareas like reinforcement learning is when researchers overlook sources of inherent stochasticity in the training process and evaluation environment~\cite{nagarajan2018deterministic,khetarpal2018re,whiteson2011protecting}.  

Other inference concerns pertain to the external validity of the functions that are learned: will they predict well on unseen data? 
In the absence of a theoretical foundation for understanding DNN performance, exploratory empirical research aims to identify proxies for properties like learnability and generalizability (e.g.,~\cite{zhang2021understanding,jiang2019fantastic}). 
Recent results show how counter to classical expectations about overfitting, minimizing training error without explicit regularization over overparameterized models tends to result in good generalization~\cite{soudry2018implicit,neyshabur2014search,zhang2021understanding}, driving a new theoretical agenda aimed at disentangling 
optimization methods and statistical properties of the solutions they find. 
Some emergent properties have been criticized. Related to shortcut learning, SGD has been shown to exhibit ``simplicity bias''---a preference for learning simple predictors first, resulting in neural nets relying exclusively on the simplest features, for example, image color and texture, and remaining invariant to complex predictive features, for example, object shape~\cite{shah2020pitfalls,kalimeris2019sgd}. 

Other concerns with external validity arise when an explicitly chosen objective function is not a good proxy for the metric of interest in using the models, called ``loss-metric misalignment'' and threatening generalization~\cite{huang2019addressing}. For example, cross-entropy loss is often used as a loss function, whereas the evaluation metric of interest is often classification error or AUCPR. More generally, reporting single scalar error measures by convention overlooks important error variation (e.g.,~\cite{drummond2006machine}).


Internal and external validity is threatened by leakage---broadly, using information from the test data in training---paralleling the reuse of data for choosing and evaluating a model's fit in psychology.
Leakage can arise through misuse of cross validation
when a single CV procedure is used for model tuning and estimating error at once~\cite{cawley2010over,hastie2009elements,hosseini2020tried}. 
Failure to carefully consider which steps involved in model training should be performed on each fold during CV can bias error estimates on test data~\cite{hastie2009elements}, as can contaminating the procedure with future data in time series applications~\cite{bergmeir2012use}. 
More generally, the use of CV for performance evaluation has been shown to lead to overoptimistic results in the presence of dependencies between the training and test set under certain conditions
~\cite{efron2004estimation,malik2020hierarchy}. 
Not unlike how low-power experiments lead to overestimates of effects in psychology, under such dependencies, underestimating test error from CV becomes more likely. 

Other issues occur in performance comparisons of models or algorithms. 
Similar to data issues in psychology, sampling error can be overlooked, including low power in performance comparisons~\cite{card2020little} and failure to acknowledge that performance estimates on the 
standard train-test splits common in benchmark datasets may not hold for randomly created train-test splits~\cite{gorman2019we}. 

Finally, implicit optimization for good performance results can also occur in ML. 
Improving performance on benchmark datasets, which have been 
thought to have caused most major ML research breakthroughs in the last 50 years~\cite{donoho201850}, is how researchers showcase improvement in model performance to get published in top conferences and journals~\cite{raji2021ai,sculley2018winner} across ML subareas (e.g., \cite{deng2009imagenet,hu2021open,wang2018glue}). This can create incentives for researchers to implicitly optimize inference around a goal of seeing their new technique rank best in performance in an evaluation, such as selectively reporting results to highlight the best accuracy achieved (Section~\ref{sec:comm}), choosing among performance measures conditional on results, or failing to acknowledge how simpler baselines perform relative to a new approach (e.g., how well the ``language prior,'' the prior distribution over labels~\cite{goyal2017making}, performs in a popular visual question answering task (VQA)~\cite{antol2015vqa}). Attempts to use OOD data to improve task performance are not valid when researchers rely on explicit knowledge of how the OOD splits were constructed or use the OOD test set for model validation~\cite{teney2020value}.

%% file: 2_4_communication.tex
\vspace{-4mm}
\section{Communication of claims}
\label{sec:comm}
Sources of error can remain unacknowledged due to communication norms that suppress uncertainty and limit reproducibility. 

\textbf{Social psychology.} 
{\em The contribution of a social psychology experiment can be framed as a stylized fact: 
a statement presumed generally true and replicable~\cite{gelman2018failure,hirschman2016stylized} about some aspect of the world. Results are used to motivate broad claims~\cite{dejesus2019generic}, with deficiencies attributable to authors failing to acknowledge exploration of multiple analysis paths contingent on the data, and tending to downplay inherent dependencies and uncertainty when describing results.}

Because stylized facts derive from the results of experiments in laboratory-like environments, often on non-rep\-re\-sen\-ta\-tive samples~\cite{gelman2018failure}, credible reporting would emphasize the specific conditions studied~\cite{simons2017constraints}. Instead, however, researchers routinely state their findings in broad terms in articles, referring to how an intervention or trait affects ``people'' or entire groups~\cite{dejesus2019generic,gutierrez2003cultural,rogoff2003cultural}, not acknowledging potential variation untested by theory and data. 

Authors can perpetuate $p$-value fallacies when they write about effects as if present or absent (e.g.,~\cite{besanccon2019continued}) or overinterpret alternative hypotheses~\cite{szucs2017null}. Or they may imply that a lack of significance is evidence of an absence of effect~\cite{altman1995statistics,besanccon2019continued} or that there is a significant difference between significant and non-significant results~\cite{gelman2006difference}.

Finally, while sharing of data and analysis code has increased in psychology in recent years, many authors have not adopted such sharing (e.g.,~\cite{tedersoo2021data}). When authors don't publish data or analysis code they used to arrive at a conclusion, readers cannot as easily identify problems or replicate the work, potentially slowing the rate at which errors that invalidate claims are caught~\cite{honesty2017,houtkoop2018data,munafo2017manifesto}.

\textbf{Machine learning.} {\em 
Communication concerns in ML include tendencies to not report trial and error over the modeling pipeline and evaluation metrics (leading to biased claims about model performance) and to downplay dependencies and uncertainty affecting performance.}

ML researchers often report point estimates of performance without quantifying uncertainty~\cite{agarwal2021deep,liao2021we,raschka2018model} or reporting key inputs such as hyperparameter and computational budget settings in non-convex optimization. This can result in performance results for which the source of empirical gains is unclear or misattributed~\cite{lipton2019research}. 
As examples, authors often do not report the number of models trained and the negative results found before the one they highlight is selected~\cite{agarwal2021deep,sculley2018winner}.
Authors may cut corners since computing uncertainty and variance in ML models can incur significant computational costs, especially for large ML models~\cite{agarwal2021deep,bouthillier2020survey}.
When not presented along with an estimate of the uncertainty of model performance arising from sources of variation like the choice of train-test split~\cite{gorman2019we}, the computational budget~\cite{dodge2019show}, the choice of hyperparameter values, and the random initialization of ML models~\cite{choi2019empirical,lucic2018gans,schmidt2021descending}, point estimates of performance represent the best-case rather than expected model performance.
Worse, researchers sometimes apply CV to tune a model then report the best performing model's error on the training set (i.e., the ``apparent error'') as if it were cross-validated error~\cite{neunhoeffer2019cross}.

As with psychology, researchers may be tempted to speculate about causes without couching them in speculative terms~\cite{lipton2019research}. 
Overgeneralization occurs from the loose connection between a task (e.g., reading comprehension, image classification) given in colloquial and anthropomorphic terms as what a model has learned to do, and a more specific definition of the problem~\cite{liao2021we,lipton2019research} for which publishable results were achieved. For example, using ``reading comprehension'' to refer to a process is misleading when the model may not have used what a human would call critical information, like the text it is ``comprehending''~\cite{kaushik2018much} (Section~\ref{sec:eval}). 
More broadly, claims about performance are rarely evaluated in the context of relevant real-world applications~\cite{liao2021we}.


ML faces analogous issues to the lack of open data and code in social psychology~\cite{dror2017replicability,gorman2019we,haibe2020transparency,sandve2013ten,stodden2014provisioning}. 
Details about dataset limitations that can threaten external validity~\cite{bender2021dangers,paullada2021data,gebru2021datasheets} are often unreported in ML literature (Section \ref{sec:data}), perhaps because new techniques for model creation have historically been valued over documenting datasets~\cite{scheuerman2021datasets}. As in psychology, checking computational reproducibility of results requires making the complete code and data available with published papers~\cite{buckheit1995wavelab}. Recent work attempts reproducibility checklists, documentation checklists, community challenges, and workshops~\cite{pineau2021improving,mitchell2019model,gebru2021datasheets}. 
However, while assessing replication in the social sciences is not trivial (e.g.,~\cite{steiner2019causal}), a somewhat unique challenge in ML is that with the creation and widespread use of large ML models requiring significant computational resources~\cite{bender2021dangers}, especially in NLP tasks, it becomes impossible for many researchers to even attempt replicating certain results. 

%% file: 3_discussion.tex
\vspace{-2mm}
\section{Implications: What can we learn from this comparison?}
As researchers move toward integrative modeling, 
they should grasp common blind spots; the summary in Table~\ref{tab:results} roughly orders issues to emphasize where concerns overlap between the two fields. 

We see evidence of different ways in which researchers place undue confidence in particular statistical methods. In ML, the use of a train/test split and cross validation can give the illusion that the inherent inability to know performance on unseen data is manageable. In social psych, belief in the power of randomized sampling and statistical testing leads researchers to overlook the importance of satisfying other assumptions or modeling other forms of variation, like in sampling stimuli. Motivating choices like model representation using asymptotic theory without considering its applicability to the specific inference problem is conventional.
In both cases, researchers' trust in methods is undergirded by unrealistic expectations about the predictability of real-world behavior and other phenomena. Social psychologists ignore the ``crud factor''~\cite{meehl1967theory} and improbability that multiple predictors thought to have large effects on the same outcome would not also correlate with one another~\cite{tosh2021piranha}. ML researchers seem to embrace the crud factor by recognizing the importance of using many predictors to avoid overfitting when the signal from any one predictor is likely to be small~\cite{dar2021farewell}, but have been slow to part with i.i.d. assumptions. 

Norms around what is publishable in each field incentivize researchers to hack results to meet implicit objectives such as statistical significance or better-than-SOTA performance, to the detriment of practical significance or external validity. 
Important dependencies in the analysis process---from types of data filtering and reuse to unacknowledged computational budgets or unspecified populations---are often overlooked, so that results do not generalize as assumed. 
Overgeneralization and suppression of uncertainty via binary statements---about the presence of effects or rank of model performance relative to baselines---are common in reporting results. 

\vspace{-2mm}
\subsection{Irrefutable claims}
On a deeper level, claims researchers are making in both fields appear to be irrefutable both by design and convention.   
In social psychology, this manifests as papers that set out to confirm hypotheses that associations will exist, or be in a certain direction, rather than mechanistic accounts that enable more specific predictions. 
When hypotheses provide only weak constraints on researchers' ability to find confirming evidence \textit{and} there is flexibility in the analysis process (not to mention incentives to publish positive evidence on often counterintuitive effects~\cite{fanelli2010positive,scargle1999publication}), ``false positive psychology''~\cite{simmons2011false} is not a surprising result. 
Consider how much more difficult, even impossible, it for those who wish to refute, rather than support, a given theory: showing no association, for example, means providing evidence for a point prediction of null effect.  
At the same time, in the absence of well-motivated stimuli sampling strategies, defined target populations, and attempts to model other sources of contextual variation, assuming that claims made about any particular parameter estimates obtained through analyzing experiment results generalize beyond that particular set of participants, stimuli, etc. is not credible. 

Turning to ML, many reproducibility failures seem to derive from a similar tolerance for irrefutable contributions, manifesting as a confusion between engineering artifacts and scientific knowledge. 
Consider a typical supervised ML paper that shows that an innovative algorithm, architecture, or model achieves some accuracy on a benchmark dataset. 
Even if we assume the reported accuracy is not optimistic for the various reasons discussed above, the researcher has contributed an engineering artifact, a tool that the practicing engineer can carry in their toolbox based on its superior performance to the state-of-the-art on a particular learning problem.  
New observations based on additional data cannot refute the performance claim of the given algorithm on the dataset, because the population from which benchmark datasets are drawn are rarely specified to the detail needed for another sample to be drawn~\cite{jo2020lessons}.
Attempts to collect a different sample from an implied population to refute claims are rare; when they have been attempted, researchers have found that the original claims no longer hold~\cite{recht2019unbiased}. Further, when researchers have tried to compare model performance across benchmark datasets, they have found that results on one benchmark rarely generalize to another, and can be fragile~\cite{torralba2011unbiased,dehghani2021benchmark}.

At a higher level, analogies between human and artificial intelligence are embedded in AI and ML culture, but are hard to render refutable. 
Without a priori specification of the neurocomputational processing involved in high-level cognition~\cite{reggia2020artificial}, 
whether ML approaches capture critical aspects of human consciousness (e.g.,~\cite{goyal2020inductive}) or new algorithms intended to instantiate human-like mechanisms (e.g.,~\cite{bengio2017consciousness}) succeed in a human-like way is entirely speculative. 

The acceptance of non-refutable research claims as research contributions, as in social psychology, creates a culture in which other methodological issues amplify the difficulty of building generalizable knowledge. 
Hubris from beliefs that big data renders modeling requirements like uncertainty quantification unnecessary~\cite{dehghani2021benchmark, bouthillier2021accounting,lin2021significant}, a lack of rigor in evaluation~\cite{sculley2018winner,liao2021we}, and overreliance on theory~\cite{lipton2019research, goldstein2022recent} may leave ML plagued with reproducibility and generalization issues. One potential bright spot lies in widespread recognition that the field is lacking foundational statistical theory to explain DNN performance. This could naturally encourage a more cautious and empirical mindset among researchers, but only if pressures to make bold claims from structural incentives that encourage ``planting one's flag'' before others do~\cite{sculley2018winner} don't outweigh the trend toward embracing uncertainty.   

\vspace{-4mm}
\subsection{Latent expectations versus reality}
Characterizing the conventions that give rise to irrefutable claims as forms of underspecification---meaning that some aspect of the learning problem has not been formalized to an extent that allows it to be solved---might help point researchers toward new methods to address what is missing. 
In particular, the role of human expectations in defining ``success'' in learning has been implicit, but innovations are often driven by making these expectations explicit.

Colloquially, many ML methods are assumed to free researchers from theorizing how well a fitted function captures critical structure in the true DGP.
Yet, many weaknesses being identified suggest that
the reality of non i.i.d. test data is pushing ML researchers in ``purely'' predictive areas toward philosophies underlying explanation. Recent definitions of underspecification~\cite{d2020underspecification}, shortcut learning~\cite{geirhos2020shortcut}, adversarial vulnerability~\cite{ilyas2019adversarial}, etc. motivate the need to impose more constraints on what is learned, and the most natural source is the human who assesses and interprets the results.

In social psychology, DGPs are modeled, if only as a symptom of using conventional inference. However, we see a displacement of prior knowledge in designing and interpreting experiments, where a priori expectations about how big an effect could be are often overlooked. There is also a failure to acknowledge how the styles of research the field rewards, such as showing that many small interventions can have large effects on a class of outcomes, are incompatible with common sense expectations of correlated effects.  

There is little reason to believe that taking steps toward integrative modeling will greatly improve practice if researchers fail to actively monitor what new methods help them learn for the issues listed in Table~\ref{tab:results}.
The worst of both worlds would result if instead they assume that any use of integrative approaches must increase rigor, because 
``now we do statistical testing,'' ``now we do human-subjects experiments,'' or ``now we use a test/train split.'' 

Another bright spot in times of methodological crisis is that when mismatch is recognized, it can lead to new technical innovations. 
Paradoxically, striving to identify a single foolproof solution to a recognized learning problem can drive new techniques to close the gap between expectations and reality. For example, in ML recognizing the ``brittleness'' of deep learning models in light of human perceptions of learning has led to major improvements to generalization from new approaches to adversarial robustness. 
ML may have an advantage over psychology in addressing reproducibility problems in that a 'cat and mouse' dynamic characterizes many recent successes, where evaluative work is followed by clever changes to the definition of a learning problem or pipeline that overcome that weakness.  
Perhaps the most important question for modern AI and ML is what, if any, forms of mismatch between human expectations and model behavior cannot be solved through a reframing of the learning problem to find beneficial new ''hacks.'' 

\vspace{-4mm}
\subsection{Epistemological gaps and rhetorical risks}
It is natural for fields to amass signals thought to be proxies of trustworthiness to enable judging work at the time of publication, when how well a claim replicates or generalizes is not known. However, a fundamental challenge in doing this is the need for reformers to recognize the incompleteness of their own knowledge.

Consider how irrefutable theories and claims induce greater dependence on imperfect ways of validating claims. In social psychology, replication is an indirect test for whether effects persist under the same or similar conditions. However, experts do not always agree on what constitutes successful replication~\cite{pawel2020sceptical}, and intuitions can be proven wrong. For example, under a formal definition of a study's reproduciblity rate, reproducing experimental results does not necessarily indicate a ``true'' effect, and vice versa for a ``false'' effect~\cite{devezer2020case}. 
In ML, tests are similarly indirect, but the stakes often higher: when an approach fails to perform as expected in the world, researchers may scrutinize the original claims, but at the expense of those affected in deployment. Progress can be hard to judge due to the speed of discovery~\cite{bowman2022dangers} and conflicting valuations of standard approaches like benchmarks (e.g.,~\cite{raji2021ai,yadav2019cold}). 


There is a need to accurately diagnose the fundamental problems, rather than symptoms only, and avoid the sort of part-for-whole substitution in reforms that drive methodological overconfidence. As fields work toward consensus views on errors, uncertainty must be embraced.
For example, debates over what core problems preregistration addresses point to the challenge of determining when a given reform should have privileged status~\cite{szollosi2019preregistration,simmons2021pre,navarro2020paths}. 



Trusting a method (whether it be a statistical idea such as Bayesian inference or causal identification, or an ML idea such as deep learning or cross validation) without examining the applied context can mislead researchers by implying that better results can be achieved via a singular universal method of statistical inference~\cite{gigerenzer2015surrogate}. It can be that more careful researchers tend to use more sophisticated methods, which will show up as a correlation between methodological sophistication and the quality of research---but this can also create an opening for methods to be used as a signal of research quality even when that is not the case.
For example, it makes sense for open-science reforms to be supported by researchers who do stronger work (and there is evidence from betting markets that experts can predict reproducibility with some accuracy~\cite{dreber2015}) and opposed by those whose work has failed to replicate (for example,~\cite{yong2012}). This would lead to open-science practices themselves being a marker of research quality. On the other hand, honesty and transparency are not enough~\cite{honesty2017}: all the openness and preregistration in the world won't endow replicability to a psychology study with a high ratio of noise to signal, which can happen with experiments whose designs focus on procedural issues (e.g., randomization), to the detriment of theory and measurement. Open-science practices can be a signal of replicability without that holding in the future. 


Steps can be taken to reduce rhetorical risks.
For example, Devezer et al.~\cite{devezer2020case} propose accompanying colloquial statements about reproducibility problems and solutions with formal problem statements and results, and provide questions to guide researchers in doing so. 
Greater rigor in reform arguments can mean quicker identification of logical errors, misintepretations of constructs, or other blind spots in attempts to steer a field back on track.